\documentclass{article}

\usepackage{PRIMEarxiv}

\usepackage[utf8]{inputenc} 
\usepackage[T1]{fontenc}    

\usepackage{nicefrac}       
\usepackage{microtype}      
\usepackage{lipsum}
\usepackage{fancyhdr}       
\usepackage{amsmath,amssymb,amsfonts}
\usepackage{textcomp}
\usepackage{xcolor}
\usepackage{tipa}
\usepackage{rotating}
\usepackage{booktabs}
\usepackage{subcaption}
\usepackage{lineno}
\usepackage{soul}
\usepackage{multirow}
\usepackage{hyperref}
\usepackage{tablefootnote}
\usepackage{xurl}
\usepackage{algorithm}
\usepackage{algpseudocode}

\usepackage{diagbox}
\usepackage{float}

\usepackage{graphicx}
\graphicspath{ {./images/} }

\title{ArtBrain: An Explainable end-to-end Toolkit for Classification and Attribution of AI-Generated Art and Style
}

\author{
  Ravidu Suien Rammuni Silva$^1$, Ahmad Lotfi$^2$,\\
  \textbf{Isibor Kennedy Ihianle$^3$, Golnaz Shahtahmassebi$^4$, Jordan J. Bird$^5$} \\
  Department of Computer Science$^{1,2,3,5}$ \\
  Department of Physics and Mathematics$^{4}$ \\
  Nottingham Trent University, United Kingdom.\\
  \texttt{$^1$ravidu.rammunisilva2024@my.ntu.ac.uk, $^2$ahmad.lotfi@ntu.ac.uk}\\
  \texttt{$^3$isibor.ihianle@ntu.ac.uk, $^4$golnaz.shahtahmassebi@ntu.ac.uk, $^5$jordan.bird@ntu.ac.uk} \\
}

\begin{document}
\maketitle

\begin{abstract}
Recently, the quality of artworks generated using Artificial Intelligence (AI) has increased significantly, resulting in growing difficulties in detecting synthetic artworks. However, limited studies have been conducted on identifying the authenticity of synthetic artworks and their source. This paper introduces AI-ArtBench, a dataset featuring 185,015 artistic images across 10 art styles. It includes 125,015 AI-generated images and 60,000 pieces of human-created artwork. This paper also outlines a method to accurately detect AI-generated images and trace them to their source model. This work proposes a novel Convolutional Neural Network model based on the ConvNeXt model called AttentionConvNeXt. AttentionConvNeXt was implemented and trained to differentiate between the source of the artwork and its style with an F1-Score of 0.869. The accuracy of attribution to the generative model reaches 0.999. To combine the scientific contributions arising from this study, a web-based application named ArtBrain was developed to enable both technical and non-technical users to interact with the model. Finally, this study presents the results of an Artistic Turing Test conducted with 50 participants. The findings reveal that humans could identify AI-generated images with an accuracy of approximately 58\%, while the model itself achieved a significantly higher accuracy of around 99\%.
\end{abstract}


\section{Introduction}
\label{sec:introduction}

A key advancement of recent generative AI models is their ability to produce visually appealing and realistic art \cite{podell_sdxl_2023,rombach_high-resolution_2022,ramesh_hierarchical_2022}. Generative AI models have become very advanced; an art generated using the model won a fine art competition meant for human artists \cite{roose_2022_an}. A recent example is the `Sony World Photography Award' \cite{grierson_2023_photographer}, which was won by an AI-generated image. Taking into account these recent developments, the future of the internet can be envisioned to be filled with synthetic artwork generated using Image Generative Models (IGMs) and other synthetic media content. The ownership and responsibility aspects of this synthetic content are still an ongoing debate and a heavily understudied topic. Due to this, the detection of synthetic data is crucial, especially for art, where due credit should always be given to the artist and the artist’s skill. 

IGMs were inspired by the ideas published in \cite{feller_theory_1949}. One of the first generative models published was Generative Adversarial Models (GAN), a type of IGM that generates images using two competing models \cite{goodfellow_generative_2014}. Variational Auto Encoder (VAE) has been explored for image generation using an Encoder-Decoder-based model setup \cite{kingma_auto-encoding_2014}. The encoder encodes the input image in a vector-based form, while the decoder uses the vector to recreate the image. GANs rapidly improved with new variants including StyleGAN \cite{karras_style-based_2019}, generating increasingly more realistic images.

Diffusion models \cite{sohl-dickstein_deep_2015} are a type of IGM that generates images by learning to destroy visual information and trying to recreate them. Later, it was improved to exceed the potentiality and capabilities of GANs \cite{ho_denoising_2020,dhariwal_diffusion_2021}. The diffusion models were based on the mathematical stochastic model called Markov Chains, which explains the probabilistic background of a linked chain of events. These types of models have significantly improved the generation quality of synthetic images, including photorealistic images \cite{saharia_photorealistic_2022} and artistic images \cite{qiao_initial_2022}. Standard Diffusion \cite{rombach_high-resolution_2022} and DALL-E \cite{ramesh_hierarchical_2022} are two of the most popular Diffusion-based Models, especially for fine art image generation. These models are also called text-to-image models because text prompts can be provided as input explaining details of the expected output image. The capabilities of diffusion models have recently become so advanced that it has become increasingly hard to differentiate between AI-generated art and human-drawn art by only using a digital image, creating a startling debate between what is real and what is not. These abilities of generative models also pose a serious issue regarding the originality of digitalised art, as authors in \cite{epstein_who_2020} ask, “Who Gets Credit for AI-Generated Art?”. 

This study does not seek to provide a definitive answer to the mentioned question but aims to offer an effective toolkit for the broader research community. This toolkit will assist in making informed decisions about artworks and help evaluate their authenticity. This toolkit includes a web-based application integrated with a model developed as a part of this study and an AI-generated artistic dataset of 10 major art styles and 180,000+ samples. Furthermore, the study compares human abilities to distinguish between human and AI-generated art with the AI's performance in this task. The main scientific contributions of this work are as follows; 

\begin{enumerate}
    \item A concise review of current research on AI-driven image generation and existing studies focused on detecting such images.
    \item Introducing AI-ArtBench - a novel AI artistic dataset, which includes samples generated using two diffusion-based models. This is the only dataset of its kind available as of this document submission.
    \item Novel CNN-based architecture, AttentionConvNeXt, based on ConvNeXt \cite{liu2022convnet} and Attention mechanism \cite{vaswani_attention_2017}.
    \item Web-based applications that can be used to verify the authenticity of artistic images via the developed models.
    \item Evaluate the human ability to identify AI-generated art images through an `Artistic Turing Test'. This is the first Artistic Turing Test conducted using complete AI art images generated by diffusion models.
\end{enumerate}

The remainder of this paper is initiated with a review of the existing work in Section \ref{sec:litreview} concisely discussing the existing status of generation, detection and attribution of AI art. Then in Sections \ref{sec:theory-and-design} and \ref{sec:implementation} the theories and techniques used in the implementation of the model and developed web-based application are explained followed by the evaluation results of the models in Section \ref{sec:evaluation}. The paper concludes with a discussion of the limitations of the current study and an exploration of future work.

\section{Related Work}
\label{sec:litreview}
This section reviews the existing work on AI art generation and detection while investigating its performance. The section also presents major datasets that can potentially be used to train models for tasks of differentiating AI art from Human art.
    

\subsection{Using AI to Analyse Art}
Many artistic styles have been identified throughout history that have been created by various artists. These are often called art movements. Art movements represent the artistic philosophies and ideas of contemporary artists who similarly saw art. Cubism \cite{gantefuhrer-trier_cubism_2004}, Surrealism \cite{nadeau_history_1965}, Impressionism \cite{herbert_impressionism_1988}, Romanticism \cite{honour_romanticism_2018}, Minimalism \cite{hornstein_understanding_2005} and Expressionism \cite{wolf_expressionism_2004} are some examples of these styles. Most of these artistic styles were pioneered in the 19th and 20th centuries, predominantly based in the European region, except for movements like Ukiyo-e \cite{harris_ukiyo-e_2012} which was formed in Japan. Datasets like \cite{strezoski_omniart_2018, mao_deepart_2017, mensink_rijksmuseum_2014, liao_artbench_2022, garcia_dataset_2020, mohammad_wikiart_2018, agapito_jenaesthetics_2015} which includes artworks with various styles mentioned above, can be used to train AI models for a wide range of tasks, from classification to artistic emotion analysis. Convolutional Neural Networks (CNNs) \cite{lecun_deep_2015} are a well-suited model for this task mainly because of their ability to extract spatial features on their own.

CNNs made image understanding through AI methods very effective. Authors LeCun et al. \cite{lecun_deep_2015} also present a method to produce heatmap-like gradient maps, which makes the results and the knowledge of CNNs explainable. Later, Selvaraju et al. \cite{selvaraju_grad-cam_2017} further improved the gradient maps making them more comprehensive. However, all these gradient map generation methods focus only on a single class. This nature of singular class representation makes it less suitable for understanding an artwork. Fused Multi-class Gradient-weighted Class Activation Map (FM-G-CAM) \cite{silva2023fmgcam} attempts to overcome this limitation by producing a saliency map that represents multiple top-predicted classes. FM-G-CAM is particularly informative compared to Grad-CAM, as it can reveal the presence of multiple artistic styles within a single painting.

\subsection{Can AI Generate art?}
The use of computer software to generate visual images was initiated long ago \cite{hertzmann_painterly_1998}. However, computers could not create art similar to human paintings until recently. This was possible due to three main factors: Large Scale Datasets, Computer Hardware Advancements, and Increased research interest in generative AI models.

Recently, deep neural networks have been used to generate painting-like images by introducing GANs \cite{goodfellow_generative_2014, zhu_unpaired_2017, karras_style-based_2019} and VAEs \cite{kingma_auto-encoding_2014}. Neural Style Transfer \cite{gatys_image_2016} presents a technique to manipulate features learnt by CNNs to transfer ‘style’ and ‘content’ from a given image to another. This could turn any image into an artwork created by a selected artist or style.

Surpassing GANs, Diffusion models \cite{sohl-dickstein_deep_2015} marked a significant shift in AI art generation. Although initial results were less impressive compared to GANs, subsequent enhancements \cite{ho_denoising_2020, dhariwal_diffusion_2021} enabled diffusion models to perform better than GANs. Diffusion models have also become particularly advanced in generating synthetic art images \cite{qiao_initial_2022}. Standard Diffusion \cite{rombach_high-resolution_2022} and DALL-E \cite{ramesh_hierarchical_2022} are two of the most popular and high-performing diffusion models. Art generated using these models is very realistic and convincing and sometimes very hard to differentiate from real human-drawn art.

Authors Song and Ermon \cite{song_generative_2019} published the Noise-Conditioned Score Network (NCSN), which was an adapted version of the original diffusion model. NCSN mainly consists of Score Matching \cite{hyvarinen_estimation_2005} and Langevin Dynamics \cite{wu_self-guided_2003}, a mathematical system for modelling molecular systems. The Denoising Diffusion Probabilistic Model (DDPM) \cite{ho_denoising_2020}, unlike NCSN, takes a series of steps, adding noise into the original images stepwise and learning how to remove the noise at each step. Then, finally, with a study published by OpenAI \cite{dhariwal_diffusion_2021}, GANs were surpassed by diffusion-based models. Rombach et al. \cite{rombach_high-resolution_2022} take a further step by conditioning the forward and backward diffusion processes using text embeddings, introducing an attention-based text-to-image diffusion model. This allowed the generation of images using textual prompts. The Mask Diffusion Model \cite{gao_masked_2023} is one of the recent studies that introduced transformer and attention-inspired variants of diffusion models. The model shows the highest FID score on the $256 \times 256$ version of ImageNet, among other generative models.

OpenAI recently published a new kind of generative model, the Consistency Model \cite{song_consistency_2023}, which was inspired by the concepts in diffusion models. Unlike diffusion models, the Consistency Model does all the steps in the forward and reverse diffusion process at once via a single-shot methodology. However, this means the model will require more parallel processing power.

Stable Diffusion XL 1.0  is another new development introduced to the original Stable Diffusion model. The new model architecture comes with a 3.5 billion parameter base model and a 6.6 billion parameter refined model that enlarges the latent image generated by the base model. Compared to the previous versions, the model boasts the capability to create very high-quality photorealism-level images with simple and shorter prompts.

Taking a completely different approach, Refik Anadol \footnote{Work of Refik Anadol avaiable here: \url{https://refikanadol.com/works/}} produces a new kind of art using generative image models and real-world data. He visualises real-world datasets in the form of art. The artworks are also a form of a dynamic Non-fungible Token (NFT), adding value to it.

\subsection{Who owns AI Art?}

Prior to verifying the authenticity of an artwork, detecting whether the given art is AI-generated is essential, to prevent misusing AI-generated art specifically in situations like competitions \cite{roose_2022_an, grierson_2023_photographer}. There are a few predominant ways of detecting AI-generated art: Analysing the metadata of the image, Comparing features derived from the image, Training a CNN to classify the real and synthetic images, and Performing a Reverse Image Search using the provided image.

In the existing literature, virtually no published study can be found that presents a method to detect AI-generated art. However, similar tasks have been carried out in a few different domains. The study \cite{rossler_faceforensics_2019} was done on the detection of synthetic human face images. The paper also presents a dataset that includes fake and real images. The study was done in three steps: Data Collection, Manipulation and Detection. Although the dataset is acceptably balanced in terms of gender, diversity factors like skin tones have not been considered for the approach. Hence, the presented accuracy might not be consistent for unseen data distributions collected from different ethnicities. The study only used synthetic images created by GANs.

The study \cite{wolter_wavelet-packets_2022} uses synthetic images generated by both GANs and diffusion-based models to create a detection mechanism for synthetic images. The mechanism involves a novel approach using wavelet transformation. The study is primarily based on detecting synthetic images created using GANs. The diffusion-generated dataset has a lower count (40k) than the GAN-generated dataset (750k). Hence, the results might not be consistent on larger, varied diffusion-generated datasets.

A recent study \cite{dogoulis_improving_2023} attempts to develop an approach to detect AI-generated images. GANs and Latent Diffusion models were used in this study for the image generation. Only higher-quality realistic images were chosen using a probabilistic quality estimation model. However, the generated images were selected only based on the quality estimation. The content of the images was not considered. Hence, the context distribution of the overall dataset could be limited.

Although many studies on deepfake detection using GANs can be found, only one major published study investigated the detection of diffusion-generated images. Since diffusion-based models have now surpassed GANs’ abilities \cite{dhariwal_diffusion_2021}, the methodologies presented in those studies are not guaranteed to work effectively for images generated by diffusion-based models. A few more unpublished preprints could also be found additionally. The dataset CIFAKE\cite{cifake_2024} holds higher importance among these. In general, very few studies can be found on detecting synthetic images. The published ones generally suffer from a low count of training data and a lack of diversity in the training data. According to paperswithcode.com (read on July 2024), only 11 papers with code could be found for Fake Image Detection\footnote{Papers with Code fake image detection paper list: \url{https://paperswithcode.com/task/fake-image-detection}}, whereas 2177 papers (read on July 2024) with code could be found for Image Generation\footnote{Papers with Code fake image generation paper list: \url{https://paperswithcode.com/task/image-generation}}. This imbalance between the research interests conveys the need for more research in AI-generated image detection and attribution.


\section{Theory and Design}

\label{sec:theory-and-design}

This section presents the theoretical background and the design of the utilised and developed components of this research. The developed components, like models, are visualised, and their mathematical nature is also explained in detail in this section.

\subsection{Data Generation using Diffusion Models}\label{diffusion-models}

To realise the research objectives, a dataset of AI-generated art was created in this study using two diffusion-based models: Standard Diffusion and Latent Diffusion \cite{rombach_high-resolution_2022}. Standard Diffusion is based on Latent Diffusion and generally produces realistic, higher-quality images. The core equation behind diffusion models is as follows:

\begin{equation}
\label{eq:x_t_step}
    x_{t} = \ \sqrt{{\overline{\alpha}}_{t}}x_{0} + \sqrt{{1 - \overline{\alpha}}_{t}}\varepsilon ,
\end{equation}

\noindent where, $x_{t}$, represents the input image at a time $t$ and ${\overline{\alpha}}_{t}$ represents the learned transformation matrix that transforms the original image $x_{0}$ into $x_{t}$. Building upon this theory, Stable Diffusion and Latent Diffusion models condition this process with textual inputs. $T$ is generally chosen as 50 in Latent Diffusion models.


The loss function $L$ used in the training process of the diffusion models is as follows:

\begin{equation}
\label{eq:diffusion-loss}
L = \mathbb{E}_{z,\epsilon\sim N(0,1),t}\left\lbrack \left\Vert \epsilon - \epsilon_{\theta}
\left( z_{t},t \right) \right\Vert_{2}^{2} \right\rbrack ,
\end{equation}

\noindent where, $z = \mathcal{E}(x)$. $\mathcal{E}$ is the encoder function that encodes the input $x$ into its latent representation $z$. $\epsilon$ represents the expected noisy image at timestep $t$ and the function $\epsilon_{\theta}\left( z_{t},t \right)$ denotes the predicted noise given the latent representation $z_{t}$ of input $x_{t}$ at timestep $t$.

\subsection{Detection Model Design}
\label{sec:model_design}
\begin{figure}[]
    \centering
    \includegraphics[width=0.95\textwidth]{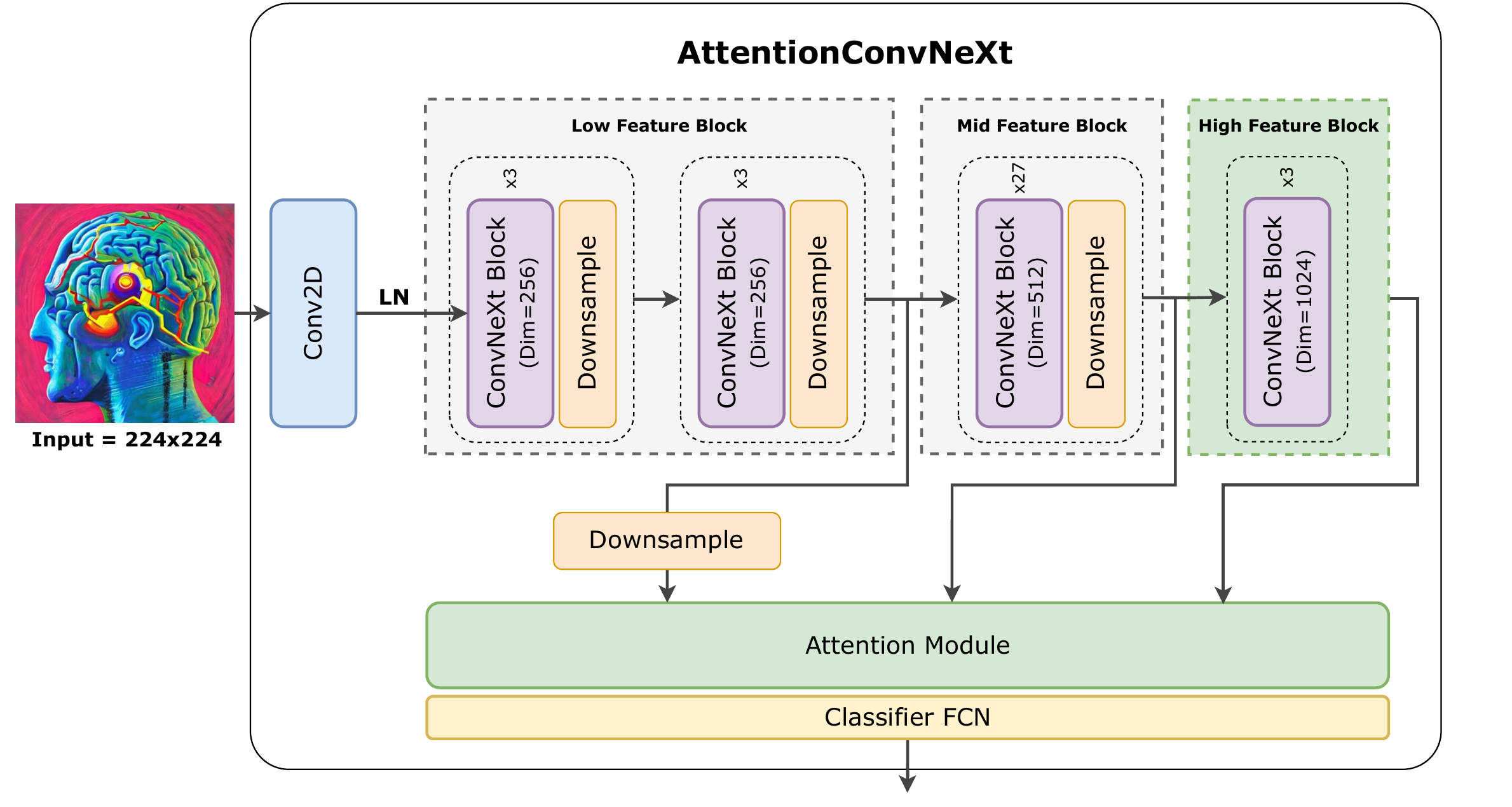}
    \caption{`AttentionConvNeXt’ model architecture design.}
    \label{fig:ArtBrain-Model}
\end{figure}

The novel model developed in this study, AttentionConvNeXt, is primarily based on Convolutional Neural Networks (CNN) \cite{lecun_deep_2015} and uses ConvNeXt model \cite{liu2022convnet} as its backbone. The model was designed to improve the ConvNeXt model by introducing an attention mechanism using the concepts of SE-Blocks \cite{hu_squeeze-and-excitation_2018}. The core idea behind the design is to get the low, mid and high-level feature maps effectively utilised for the final prediction. Figure \ref{fig:ArtBrain-Model} shows the high-level model architecture of the AttentionConvNeXt model.

`Low feature block' and `Mid feature block' are pre-trained using ImageNet \cite{deng_imagenet_2009} dataset and are frozen during the model's training process retaining low-level features extracted from the ImageNet dataset. Unlike Low and Mid-feature blocks, the `High feature block' is trainable and aims to extract higher-level features of the training images. This block was initially loaded with the weights trained on ImageNet and further fine-tuned in training. `Attention Module' receives inputs from all the feature blocks mentioned above where all the blocks are concatenated into one deep feature map block, which is then weighted using the concept of `squeeze and excitation' \cite{hu_squeeze-and-excitation_2018}. `Classifier FCN' receives the weighted feature maps from the Attention module, which are then used to classify the targeted inputs via a fully connected neural network that outputs a probability distribution.

\subsection{Attention Module}
\label{attention-module}

\begin{figure}[]
    \centering
    \includegraphics[width=0.8\textwidth]{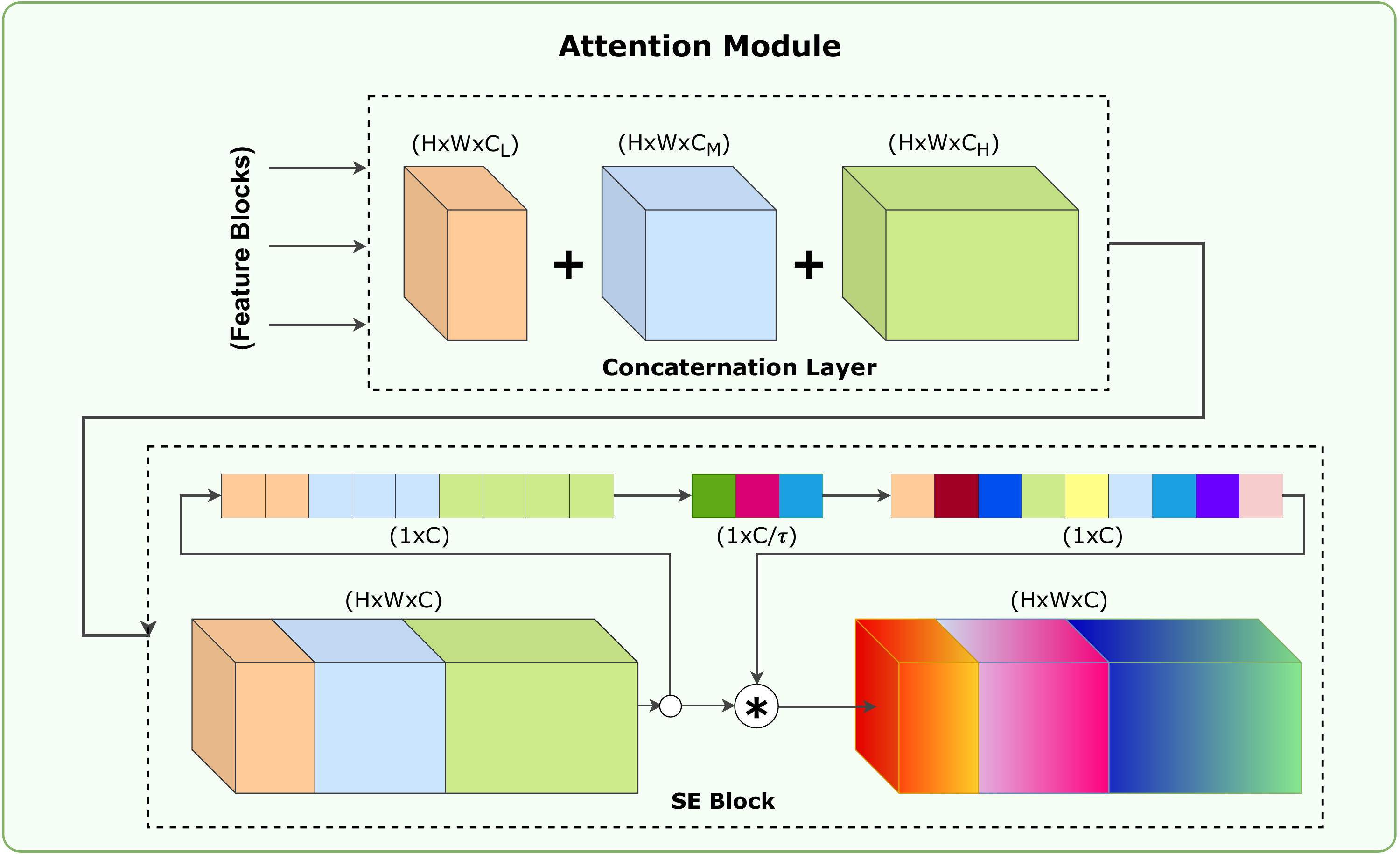}
    \caption{Attention Module Architecture.}
    \label{fig:attention-module-arch}
\end{figure}

The Attention Module in Figure \ref{fig:attention-module-arch} uses the concept of SE-Block \cite{hu_squeeze-and-excitation_2018} and re-purposes it similarly to the author's research ParallelXNet \cite{rammuni_silva_effective_2022}. Instead of a single block of feature maps, it focuses on multiple blocks of feature maps taken from different depths of the backbone CNN network, ensuring that `attention' is given to the mid-feature and end-feature maps. $H \times W \times C$ is the size of the concatenated block $Z_{concat}$ where $C = C_{low} + C_{mid} + C_{high}$. $C_{low}, C_{mid}, C_{high}$ respectively represents the channel count of the low, mid, and high feature blocks inputted into the Attention Module, as explained in Detection Model Design Section \ref{sec:model_design}:

\begin{equation}
    Z_{concat} \in \mathbb{R}^{H \times W \times C} ,
\end{equation}

\begin{equation}
    y_{c} = \frac{1}{I \times J}\sum_{i}^{I}{\sum_{j}^{J}z_{i,j,c}},\ \ where\ y_{c} \in \ \mathbb{R}^{I \times J} ,
\end{equation}

\begin{equation}
    Y = \begin{bmatrix}
    y_{1} & \ldots & y_{C}
    \end{bmatrix}^{T},\ \ where\ Y \in \mathbb{R}^{I \times J \times C} ,    
\end{equation}

\noindent where, $y_{c}$, represents the averaged value for each channel $c$. Then, the channel importance, $\omega$, is calculated as in the Equation \ref{eq:final-sgmoid} by passing $Y$ through two fully connected layers where training parameters, $W_{1}$ and $W_{2}$, respectively:

\begin{equation}
\label{eq:final-sgmoid}
    \omega = \ sigmoid\left\lbrack W_{2} \left( ReLU\left( W_{1} Y \right) \right) \right\rbrack,\ \ where\ \omega \in \mathbb{R}^{1 \times C} .
\end{equation}

In Equation \ref{eq:final-sgmoid}, $W_{1}\mathbb{\in R}^{\frac{C}{\mu} \times C},\ W_{2}\mathbb{\in R}^{C \times \frac{C}{\mu}}$, where, $\mu$, is the feature reduction factor that `squeezes' the feature importance matrix that exists in the first FCN of the SE-Block. Then, the channel importance is used to weight the feature maps (channels) of the original feature map block as in Equation \ref{eq:z-weighting}. Note that the FCN layers here do not include a bias parameter:

\begin{equation}
\label{eq:z-weighting}
    Z^{c} = \left( \omega^{c} \right)\left( Z_{concat}^{c} \right),\ \ where\ Z^{c} \in \mathbb{R}^{I \times J} ,
\end{equation}

\begin{equation}
    Z_{weighted} = {\lbrack\begin{matrix}
    Z^{1} & \ldots & Z^{C}
    \end{matrix}\rbrack}^{T},\ \ where\ Z_{weighted} \in \mathbb{R}^{I \times J \times C} ,
\end{equation}

\noindent $Z_{weighted}$ represents the final output of the Attention Module, which is then connected to the Classifier Block.

\section{Implementation}
\label{sec:implementation}

This section discusses the implementation decisions and the techniques used in developing the prototypes related to this study including the novel AI-art dataset. The three predominant implementation tasks carried out in this project are: Data Generation, CNN Model Development, and ArtBrain Application development.

\subsection{Novel Dataset: AI-ArtBench}\label{subsec:novel-dataset}

\begin{figure}[]
    \centering
    \includegraphics[width=0.95\textwidth]{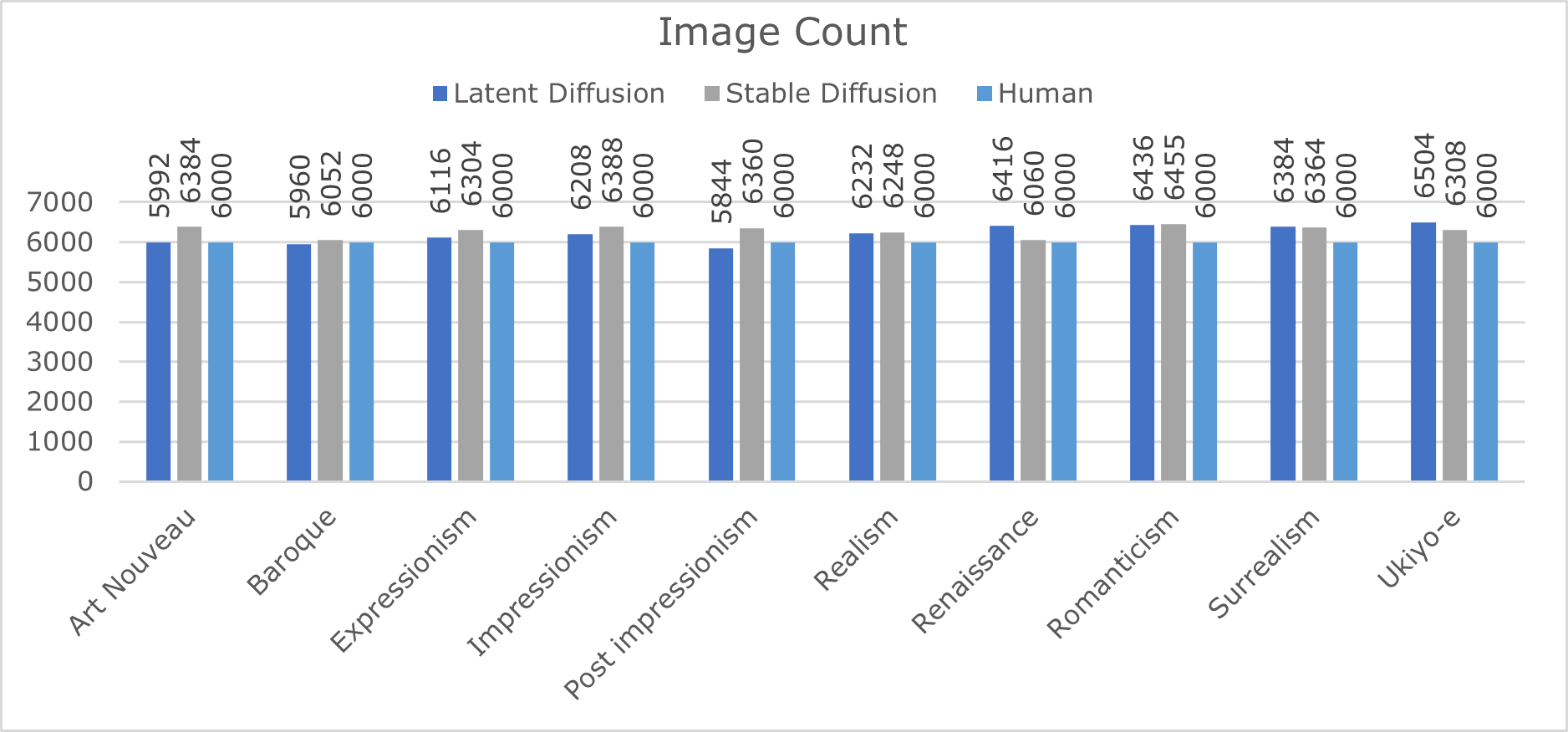}
    \caption{Image counts in each class.}
    \label{fig:data-stats}
\end{figure}

\begin{figure}[]
    \centering
    \includegraphics[width=0.95\textwidth]{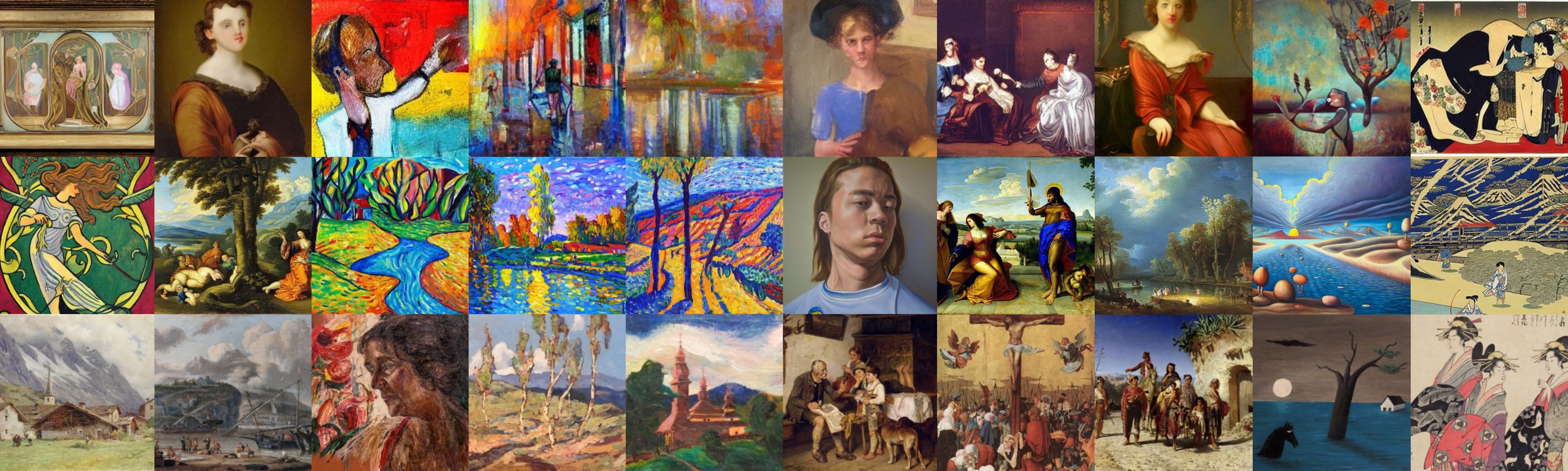}
    \caption{Sample of the dataset representing each style generated using each source. Rows (top to bottom): Latent Diffusion, Standard Diffusion, Human. Columns (left to right): Art Nouveau, Baroque, Expressionism, Impressionism, Post impressionism, Realism, Renaissance, Romanticism, Surrealism, Ukiyo-e.}
    \label{fig:collage-dataset}
\end{figure}

This study presents a novel dataset containing AI-generated art images related to 10 major art styles: AI-ArtBench. The samples were created using two main diffusion-based models with official pre-trained weights: Latent Diffusion\footnote{Latent Diffusion model repository: \url{https://github.com/CompVis/latent-diffusion}} and Stable Diffusion\footnote{Standard Diffusion model repository: \url{https://github.com/CompVis/stable-diffusion}}. Using this, together with the real art dataset ArtBench\footnote{Original ArtBench dataset can be accessed here: \url{https://artbench.eecs.berkeley.edu/files/artbench-10-imagefolder-split.tar}}, a new dataset of 185,015 images, including 125,015 AI-generated images, AI-ArtBench\footnote{AI-ArtBench dataset can be accessed here: \url{https://www.kaggle.com/datasets/ravidussilva/real-ai-art}}, was compiled. The summary of the compiled dataset is provided in Table \ref{tab:dataset-info} and Figure \ref{fig:data-stats}. The dataset is split into two splits: Train and Test. Train split is to be used for the training purposes of the model, whereas the Test split is for evaluation purposes. The testing split is perfectly balanced even among the subclasses under each source, allowing it to be used as an unbiased validation set in evaluating the models. Each split is divided into 30 sub-classes representing the ten art styles for each source. Appendix B contains the image count per each class. A sample from the dataset is displayed in Figure \ref{fig:collage-dataset}.

\begin{table}[]
\centering
\caption{Data count from each source.}
\label{tab:dataset-info}
\resizebox{0.6\textwidth}{!}{%
\begin{tabular}{lllll}
\hline
\multirow{2}{*}{\textbf{Source}} & \multirow{2}{*}{\textbf{Image Size}} & \multirow{2}{*}{\textbf{Class Count}} & \multicolumn{2}{l}{\textbf{Count}}          \\ \cline{4-5} 
                 &         &    & Train  & Test   \\ \hline
Latent Diffusion & $256 \times 256$ & 10 & 52,092 & 10,000 \\
Stable Diffusion & $768 \times 768$ & 10 & 52,923 & 10,000 \\
Human            & $256 \times 256$ & 10 & 50,000 & 10,000 \\ \hline
\multicolumn{2}{l}{\textbf{Total}}                                      & \textbf{30}                           & \textbf{155,015} & \textbf{30,000} \\ \hline
\end{tabular}%
}
\end{table}\textbf{}

The following sections contain the configurations, prompts, and parameters set in the image generation for each model.

\paragraph{Latent Diffusion}\label{latent-diffusion}

The same prompt was used for generating all the art images, treating all the artistic styles equally. No further conditioning was added to the prompts, allowing the generated images to be generalisable. Each sample generation iteration is done using a generated random seed between 0 and 999999999, and the seed is stored in the filenames of the images. 
Configurations of the Latent diffusion model used to generate the images
are as follows: Prompt: {[}``A painting in \textless art style\textgreater{} art style"{]}, Model Configuration: [Diffusion Steps - 50, Image Size - $256 \times 256$ \textbar{} Type: JPEG, Parallel Samples - 4, Diversity Scale - 5.0, CLIP Model - ViT-B-32, Sampler - PLMS]

\paragraph{Stable Diffusion}\label{stable-diffusion}

Similarly to the Latent Diffusion model, the same prompt was used for generating all the art images, treating all the artistic styles equally. However, since Stable Diffusion v2-1\footnote{Used Stable diffusion model available here: \url{https://huggingface.co/stabilityai/stable-diffusion-2-1}} allows inputting a negative prompt, this was used to remove the photo frame that was observed in most of the generated AI art images in the pre-tests. xFormers\footnote{Code for xFormers available here: \url{https://github.com/facebookresearch/xformers}} was used for efficient image generation. Each sample generation is done using a generated random seed between 0 and 999999999, and the seed is stored in the filenames of the images. An identical prompt to the latent diffusion model was used with a the following Negative Prompt: {[}``photo frame''{]}. Model configurations were set as follows: [Diffusion Steps - 50, Image Size: 768x768 \textbar{} Type: JPEG, Parallel Samples - 4, Guidance Scale - 9, Sampler - DPMS Multistep Scheduler]

Intel Xenon 2.0 GHz with 13 GB of RAM and Nvidia Tesla T4 with 16GB VRAM were used for the sample generation. Samples were named as: [\textless class index\textgreater-\textless generation seed\textgreater-\textless random number\textgreater.jpg]

\subsection{CNN Model Development}\label{cnn-model-development}

Mainly, two CNN models were implemented and trained in this study: AttentionConvNeXt, explained in Section \ref{sec:model_design} and MobileNetV2 introduced in \cite{sandler_mobilenetv2_2018}. AttentionConvNeXt was developed, focussing on accuracy, and MobileNetV2 was developed, focussing on performance and on-device ML. Kaggle Notebooks with Intel Xenon 2.0 GHz, 13 GB RAM, and Nvidia Tesla P100 GPU with a VRAM of 16 GB were used for all the model training.

AttentionConvNeXt is developed using ConvNeXt as the primary network. The following sections contain technologies used, training configurations, and other technical details of the model.

\paragraph{Preprocessing}\label{preprocessing}

\begin{figure}[]
    \centering
    \includegraphics[width=0.65\textwidth]{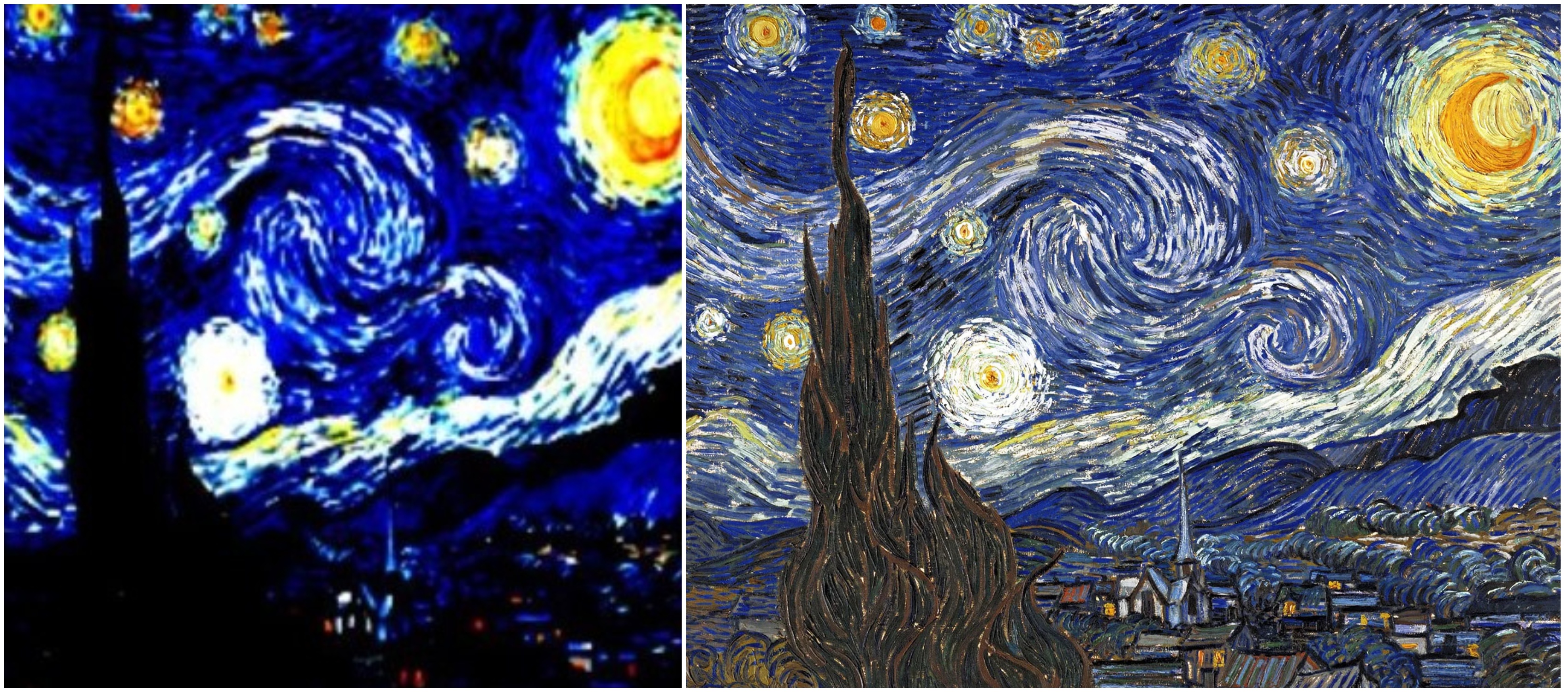}
    \caption{Preprocessed image Vs Original image (Starry Night by Vincent-van-Gogh, 1889).}
    \label{fig:preprocess-comparison}
\end{figure}

The images of the compiled dataset were preprocessed according to the parameters
of the ImageNet dataset\cite{deng_imagenet_2009}. Figure \ref{fig:preprocess-comparison} compares the visual appearance after preprocessing the image.

\paragraph{Model Development and training}
\label{sec:model-development-and-training}

The explained model in Section \ref{sec:model_design} was developed using PyTorch and the code is available open-source\footnote{Model code available here: \url{https://github.com/SuienS/ai-art-detector/tree/master/AppFastAPI/model}}. PyTorch's recommended object-oriented design was followed in all the implementations. The implementation was done in two PyTorch modules: AttentionModule (Section \ref{attention-module}) and AttentionConvNeXt (Section \ref{sec:model_design}). Firstly, all the feature map blocks were concatenated along the channel axis. Then, the block was traversed through an FCN, retrieving the channel importance. Then, it was used to weight the original feature map blocks in AttentionConvNeXt. Then, the weighted new feature block is inputted to the classifier FCN. Classifier FCN mainly contains two linear layers, with dropout layers in between to prevent overfitting. Hyperparameters were selected after a series of optimisation experiments. Batch size was set to 32 and the model was trained for 18 epochs until the learning plateaus. Learning rate was set to 0.001 at the beginning and reduced by factor of 10 with a patience of 2 epochs with the Adam optimiser.


As explained in Section \ref{sec:model_design}, the model was preloaded with ImageNet weights, and the low and mid feature extraction layers were frozen. The testing set was used as the Validation set in the training. The learning rate was scheduled to reduce by a factor of 10 whenever two consecutive increments in the validation loss related to adjacent epochs were reported. After training for 18 epochs, weights that caused the lowest validation loss for an epoch, the 15\textsuperscript{th} Epoch, was selected. Validation loss was selected rather than the validation accuracy since the focus was on the `overall' model's correctness rather than just the ability to predict the correct class. This also makes the activation maps used to develop FM-G-CAM \cite{silva2023fmgcam} as reliable as possible. The model was then deployed using FastAPI\footnote{FastAPI application code available here: \url{https://github.com/SuienS/ai-art-detector/tree/master/AppFastAPI}}.

\paragraph{On-device ML: MobileNet V2}\label{on-device-ml-mobilenet-v2}
Another lite model was developed to allow the users to use the model on-device. The focus of this model was more on efficiency than accuracy. Data preprocessing and training were performed identically to the AttentionConvNeXt. However, TensorFlow was used to develop and train this model for better support. Then the trained model was saved in JSON format, making it possible to host it using TensorFlowJS\footnote{TFJS application code available here: \url{https://github.com/SuienS/ai-art-detector/tree/master/AppTFJS}}. Further, a standalone HTML was developed embedding all the dependencies required to run the application having the model alongside\footnote{Standalone HTML application code available here: \url{https://github.com/SuienS/ai-art-detector/tree/master/AppTFJS/standaloneHTML}}. The trained model was deployed was using TensorFlowJS via JavaScript, allowing offline usage via a web browser on-device.


\section{Evaluation}
\label{sec:evaluation}

This section evaluates all the prototypes built as part of this study. In this section, the accuracy of the developed models will be evaluated and compared with the current state-of-the-art. A consistent testing set was used for all the tests conducted in this section. The same set used for validation is used as the test set. However, the batch size was set to 1 in the inferencing to ensure the tests simulate the real-world environment. Each sample was preprocessed identically to the training introduced in Section \ref{preprocessing}. Accuracy is measured in terms of both the art style and the source model. F1-Score was used as a measure of accuracy for the classes.

\subsection{Model Accuracy: Classification}\label{classification-accuracy}

Table \ref{tab:classification-scores} presents the classification accuracy of the trained models comparatively to the current state-of-the-art. The proposed approach maintains the best F1-score for 28 classes out of 30, making it the most accurate model. MobileNet V2 model also maintains a reasonable accuracy of 84\% while having the ability to process samples comparatively faster. Regarding class F1-scores, Human-Realism has the lowest score, and SD-Ukiyo-e has the highest score in both the trained models. This could be due to the significant visual differences in Ukiyo-e style and the variety of the visual structures exist in Human-Realism style.

\begin{table}[]
\centering
\caption{Classification scores.}
\label{tab:classification-scores}
\resizebox{0.75\textwidth}{!}{%
\begin{tabular}{lllll}
\cline{3-5}
                            &                    & \multicolumn{3}{l}{\textbf{F1-Score}} \\ \hline
\textbf{Generative Model}    & \textbf{Art Style}   & \textbf{ArtBench} & \textbf{MobileNet V2} & \textbf{Our Study} \\ \hline
\textbf{Latent Diffusion}   & Art Nouveau        & -           & 0.9685     & 0.9875     \\
\textbf{}                   & Baroque            & -           & 0.9243     & 0.9565     \\
\textbf{}                   & Expressionism      & -           & 0.9667     & 0.9766     \\
\textbf{}                   & Impressionism      & -           & 0.7238     & 0.7627     \\
\textbf{}                   & Post impressionism & -           & 0.7247     & 0.7551     \\
\textbf{}                   & Realism            & -           & 0.8808     & 0.9099     \\
\textbf{}                   & Renaissance        & -           & 0.9332     & 0.9542     \\
\textbf{}                   & Romanticism        & -           & 0.8906     & 0.9190     \\
\textbf{}                   & Surrealism         & -           & 0.9800     & 0.9885     \\
\textbf{}                   & Ukiyo-e            & -           & 0.9995     & 1.0000     \\ \hline
\textbf{Standard Diffusion} & Art Nouveau        & -           & 0.9990     & 1.0000     \\
                            & Baroque            & -           & 0.9861     & 0.9935     \\
                            & Expressionism      & -           & 0.9975     & 0.9985     \\
                            & Impressionism      & -           & 0.9885     & 0.9935     \\
                            & Post impressionism & -           & 0.9870     & 0.9935     \\
                            & Realism            & -           & 0.9975     & 1.0000     \\
                            & Renaissance        & -           & 0.9874     & 0.994      \\
                            & Romanticism        & -           & 0.9975     & 0.9985     \\
                            & Surrealism         & -           & 0.9985     & 1.0000     \\
                            & Ukiyo-e            & -           & 1.0000     & 1.0000     \\ \hline
\textbf{Human}              & Art Nouveau        & 0.6660      & 0.6209     & 0.7043     \\
                            & Baroque            & 0.7917      & 0.6948     & 0.7854     \\
                            & Expressionism      & 0.5127      & 0.5122     & 0.5881     \\
                            & Impressionism      & 0.4636      & 0.4728     & 0.5228     \\
                            & Post impressionism & 0.5175      & 0.4660     & 0.5759     \\
                            & Realism            & 0.4525      & 0.4079     & 0.4991     \\
                            & Renaissance        & 0.8159      & 0.7314     & 0.8111     \\
                            & Romanticism        & 0.5980      & 0.4780     & 0.6050     \\
                            & Surrealism         & 0.7737      & 0.7618     & 0.8251     \\
                            & Ukiyo-e            & 0.9695      & 0.9721     & 0.9850     \\ \hline
\multicolumn{2}{l}{\textbf{Overall Model Accuracy}} & \textbf{-}        & \textbf{0.8354}       & \textbf{0.8693}    \\ \hline
\end{tabular}%
}
\end{table}

\subsection{Model Accuracy: Attribution}\label{attribution-accuracy}

Table \ref{tab:attribution-scores} presents the attribution accuracy of the trained models. Attribution scores of main classes are derived by averaging their subclasses' prediction scores. The scores represent how well the models can attribute the images generated using diffusion models. Both models are highly accurate in attribution.

\begin{table}[]
\centering
\caption{Attribution accuracy.}
\label{tab:attribution-scores}
\resizebox{0.55\textwidth}{!}{%
\begin{tabular}{lll}
\cline{2-3}
                                & \multicolumn{2}{l}{\textbf{F1-Score}} \\ \cline{2-3} 
\textbf{Generative Model} & \textbf{MobileNet V2} & \textbf{\begin{tabular}[c]{@{}l@{}}Our Study  \\ (A-ConvNeXt)\end{tabular}} \\ \hline
Latent Diffusion                & 0.9996            & 0.9997            \\
Standard Diffusion              & 0.9999            & 1.0000            \\
Human                           & 0.9995            & 0.9996            \\ \hline
\textbf{Overall Model Accuracy} & \textbf{0.9997}   & \textbf{0.9997}   \\ \hline
\end{tabular}%
}
\end{table}

\subsection{Ablation Tests}\label{ablation-tests}

A series of ablation tests were conducted to measure the effectiveness of the techniques used. The results in Table \ref{tab:ablation-results} shows that using ImageNet weights significantly improves the model\textquotesingle s overall accuracy. This also implies the requirement for more training data. The newly introduced architecture AttentionConvNeXt increases the accuracy by around 3\% overall. It also maintains the highest F1-score on 29 of 30 total classes across all three models. Further, AttentionConvNeXt with transfer learning scores 10\% better in some classes when compared to ConvNeXt with transfer learning. The accuracy improvement shows the benefit of using the Attention Module in the AttentionConvNeXt Model.

\begin{table}[]
\centering
\caption{Ablation test results.}
\label{tab:ablation-results}
\resizebox{0.8\textwidth}{!}{%
\begin{tabular}{lllll}
\cline{3-5}
                   &                    & \multicolumn{3}{l}{F1-Score} \\ \cline{1-5} 
Generative Model &
  Art Style &
  ConvNeXt &
  ConvNeXt + TL &
  \begin{tabular}[c]{@{}l@{}}Our Approach\\ (A-ConvNeXt + TL)\end{tabular} \\ \hline
Latent Diffusion   & Art Nouveau        & 0.9337   & 0.9707  & 0.9875  \\
                   & Baroque            & 0.8639   & 0.9331  & 0.9565  \\
                   & Expressionism      & 0.9071   & 0.9557  & 0.9766  \\
                   & Impressionism      & 0.7100   & 0.7351  & 0.7627  \\
                   & Post impressionism & 0.6575   & 0.6651  & 0.7551  \\
                   & Realism            & 0.8053   & 0.8714  & 0.9099  \\
                   & Renaissance        & 0.8751   & 0.9337  & 0.9542  \\
                   & Romanticism        & 0.8138   & 0.8803  & 0.9190  \\
                   & Surrealism         & 0.9226   & 0.9701  & 0.9885  \\
                   & Ukiyo-e            & 0.9872   & 1.0000  & 1.0000  \\ \hline
Standard Diffusion & Art Nouveau        & 0.9910   & 0.9965  & 1.0000  \\
                   & Baroque            & 0.9615   & 0.9935  & 0.9935  \\
                   & Expressionism      & 0.9842   & 0.9960  & 0.9985  \\
                   & Impressionism      & 0.9628   & 0.9904  & 0.9935  \\
                   & Post impressionism & 0.9453   & 0.9866  & 0.9935  \\
                   & Realism            & 0.9748   & 0.9980  & 1.0000  \\
                   & Renaissance        & 0.9711   & 0.9950  & 0.9940  \\
                   & Romanticism        & 0.9828   & 0.9980  & 0.9985  \\
                   & Surrealism         & 0.9875   & 0.9960  & 1.0000  \\
                   & Ukiyo-e            & 0.9916   & 1.0000  & 1.0000  \\ \hline
Human              & Art Nouveau        & 0.4672   & 0.6494  & 0.7043  \\
                   & Baroque            & 0.5858   & 0.7341  & 0.7854  \\
                   & Expressionism      & 0.3150   & 0.5301  & 0.5881  \\
                   & Impressionism      & 0.3928   & 0.4700  & 0.5228  \\
                   & Post impressionism & 0.3871   & 0.5271  & 0.5759  \\
                   & Realism            & 0.3474   & 0.4493  & 0.4991  \\
                   & Renaissance        & 0.5428   & 0.7588  & 0.8111  \\
                   & Romanticism        & 0.3538   & 0.5659  & 0.6050  \\
                   & Surrealism         & 0.5446   & 0.7926  & 0.8251  \\
                   & Ukiyo-e            & 0.8674   & 0.9762  & 0.9850  \\ \hline
\multicolumn{2}{l}{Overall Model Accuracy} &
  0.7717 &
  0.8450 &
  0.8693 \\ \hline
\end{tabular}%
}
\end{table}

\subsection{Quantitative Evaluation}
\label{sec:quantitative-eval}

This section evaluates the system based on the qualitative factors related to AI art detection and attribution.

\subsubsection{ArtBrain on popular artworks}\label{artbrain-on-popular-artworks}

The following section presents the ArtBrain model prediction results on a few artworks the model has never seen.

``Mona Lisa" by Leonardo Da Vinci is one of the most famous artworks ever created by an artist shown in Figure \ref{fig:monalisa-pred} of the Appendix. It was created in 1503 and is a well-known Renaissance-style artwork, and the model accurately detects it as `Renaissance - Human' with 85\% confidence. Also, as correctly shown in the heatmap, a few spots of the image show Baroque-style visual structures. ``Starry Night" by post-impressionist artist Vincent Van Gogh is used in Figure \ref{fig:starrynight-pred} of the Appendix. The model correctly identified it, and some features that bring expressionist characteristics were also accurately highlighted.

A controversial artwork by Jason Michael Allen called ``Space Opera'', which won an art competition\footnote{News article can be accessed here: \url{https://www.nytimes.com/2022/09/02/technology/ai-artificial-intelligence-artists.html}} that was meant for human artists, was used on the ArtBrain model as in Figure \ref{fig:spaceOpera-pred} of the Appendix. The model identifies it as Romanticism art created using Standard Diffusion. The art used here was generated using Midjourney\footnote{Midjourney official site: \url{https://www.midjourney.com/}}, an application with a Diffusion-based model derived from Stable Diffusion. Hence, the ArtBrain model is successful in this scenario as well. This provides an indication of the models performance against the unseen data distributions. However, further research is needed on the capability of the model on identifying samples generated from generative models other than the ones used in the training.

\subsubsection{Effect of image contrast on the prediction}\label{effect-of-image-contrast-on-the-prediction}

\begin{figure}[]
    \centering
    \includegraphics[width=0.85\textwidth]{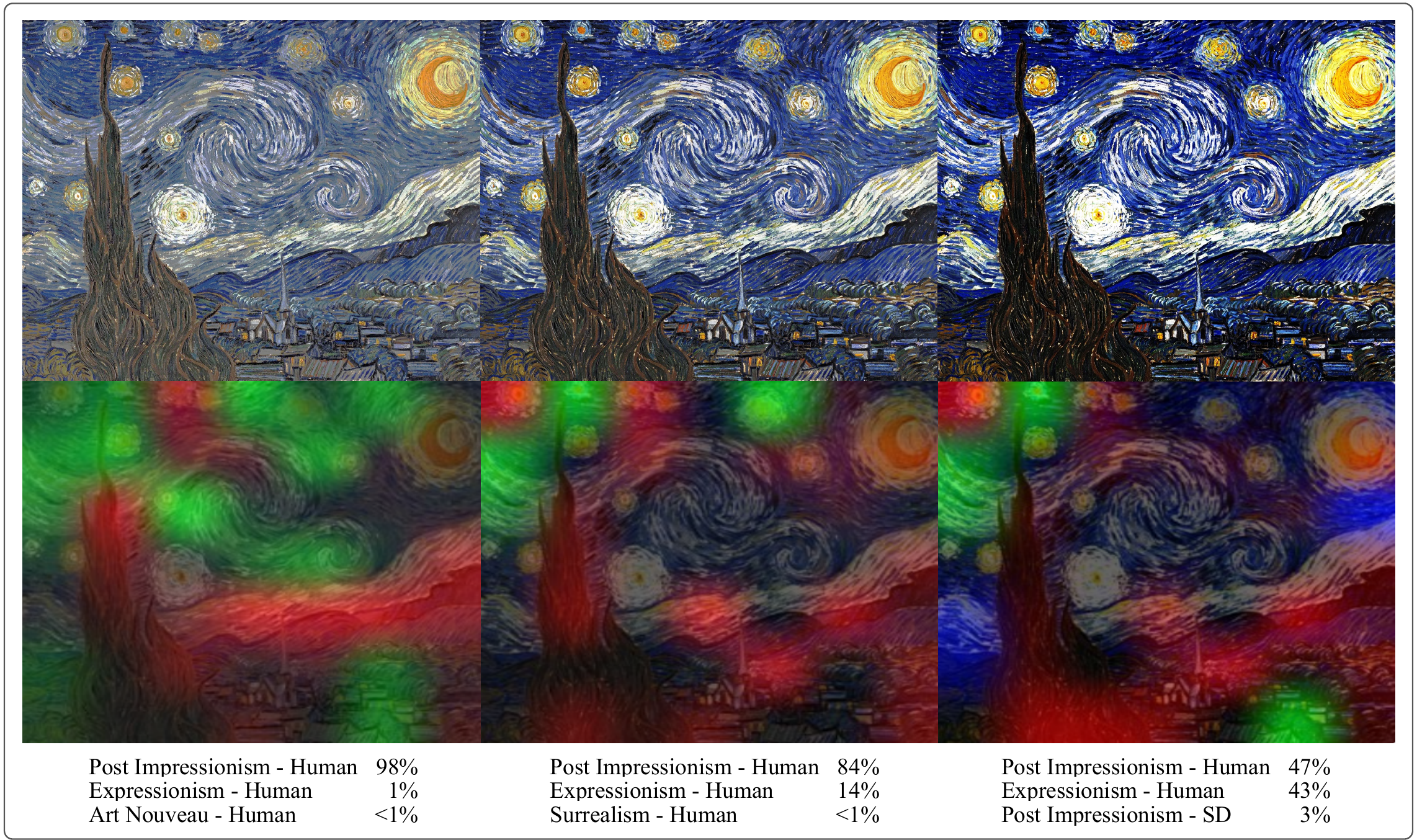}
    \caption{Prediction results on ``Starry Nights" with different contrast levels. Left: -100\% Contrast, Middle: Reference, Right: +100\% Contrast.}
    \label{fig:contrasteffext-pic}
\end{figure}

In this test, as in Figure \ref{fig:contrasteffext-pic}, the parameters of digital images were investigated on their the effect to the model predictions. The chosen parameter was the `Contrast' of the image and it can be concluded that changes to the contrast can significantly impact the model predictions.

\subsubsection{Predictions on replicated images}\label{predictions-on-replicated-images}

\begin{figure}[]
    \centering
    \includegraphics[width=0.55\textwidth]{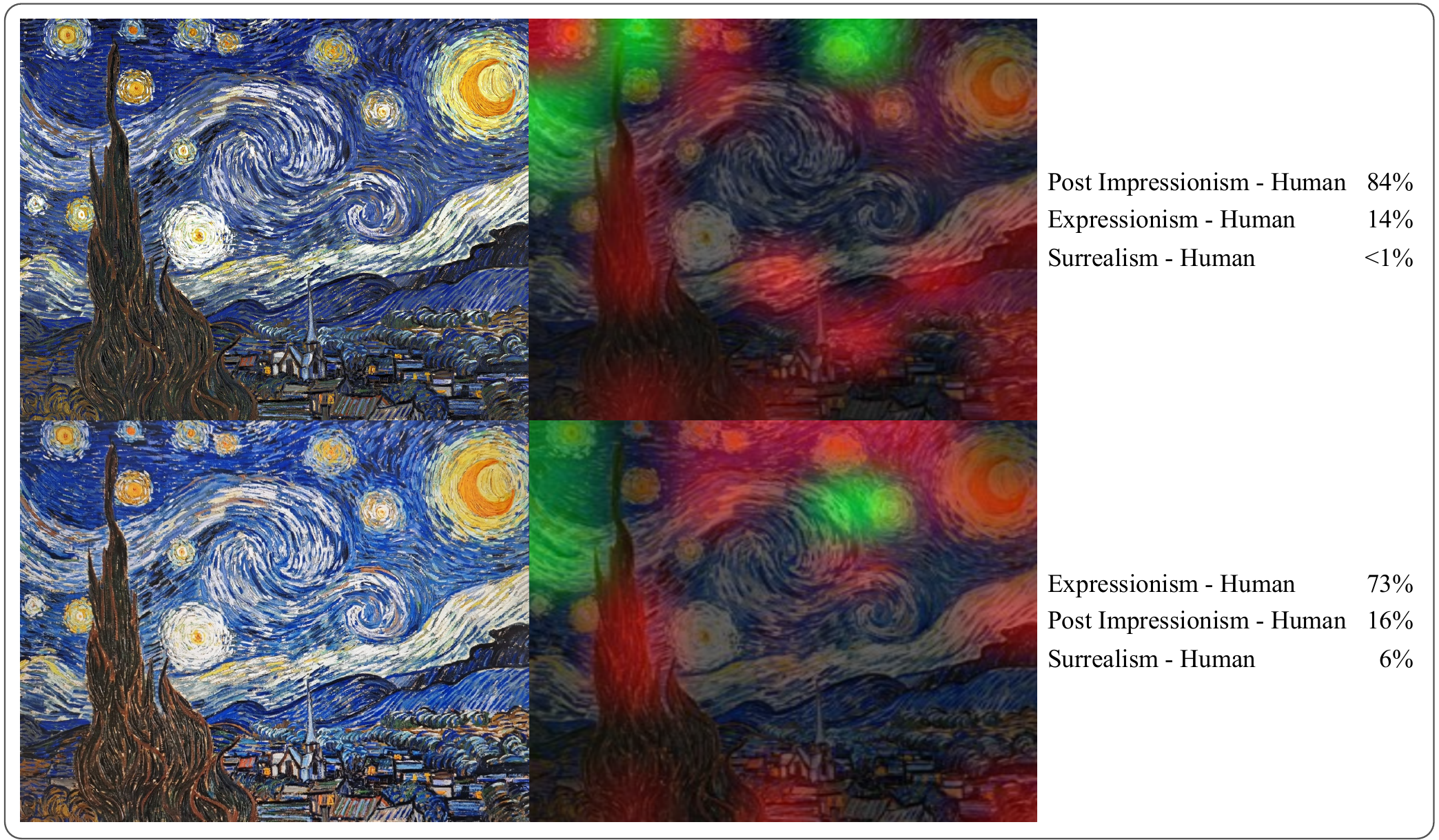}
    \caption{Prediction results of ``Starry Nights" (top row) and its replication (bottom row).}
    \label{fig:replicated-pic}
\end{figure}

For this test, as in Figure \ref{fig:replicated-pic}, ``Starry Nights" is compared to its replicated version. The replicated version on the right was created by capturing the original version of the image displayed on a 2K laptop screen. The capture was taken with a Samsung Galaxy Note 20 mobile phone with a 1.8f aperture, 9MP quality, and ISO 200 conditions. As it is evident from the prediction results, even though both images are very similar to those of the human eye, the prediction results have changed drastically.

\subsection{Artistic Turing Test}\label{artistic-turing-test}

\begin{figure}[]
    \centering
    \includegraphics[width=0.65\textwidth]{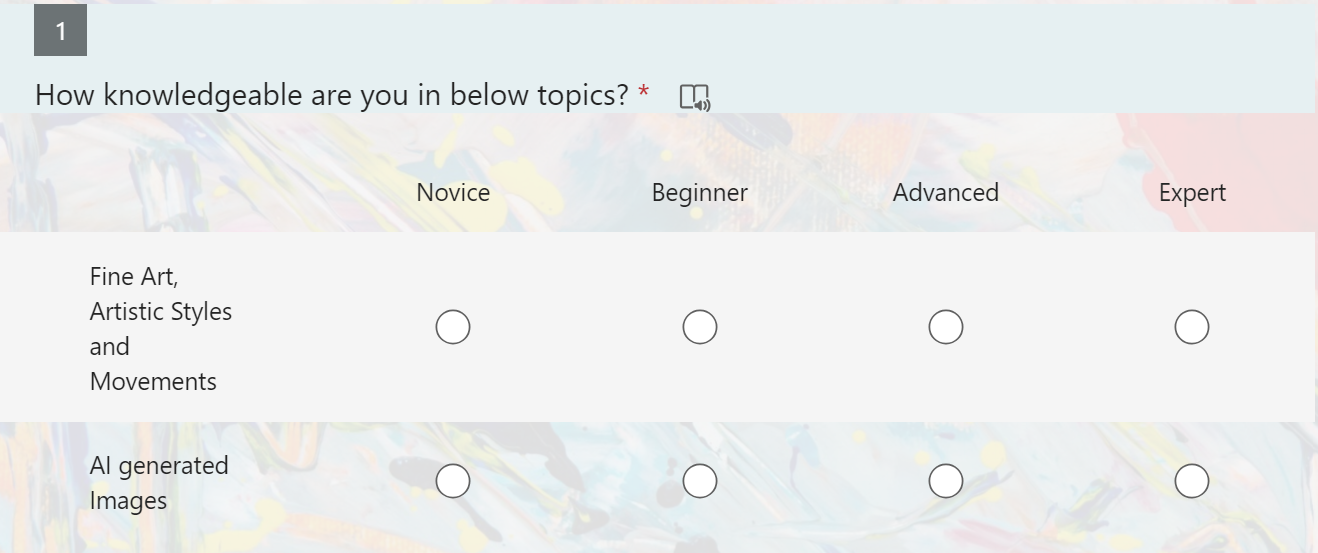}
    \caption{Initial Question of `Artistic Turing Test'.}
    \label{fig:turingtest-initial}
\end{figure}

\begin{figure}[]
    \centering
    \includegraphics[width=0.85\textwidth]{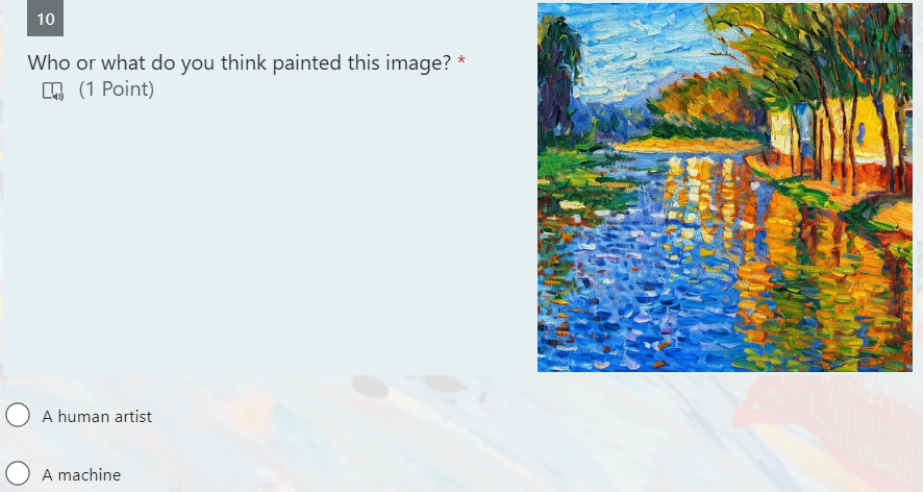}
    \caption{‘Artistic Turing Test’ Question example. This is an AI-generated art which 84\% of the Artistic Turing Test participants wrongly identified as a Human art.}
    \label{fig:turingtest-qs}
\end{figure}

In this study, ``Artistic Turing Test" was conducted resting the practical effectiveness of the system. The test was based on the original ``Turing Test" \cite{turing2009computing}, built to evaluate general machine intelligence. Authors Daniele et al. \cite{romero_what_2021} have performed a similar test but only using brush strokes rather than whole images. This test in this study is the first test conducted using diffusion-generated full art images. 


In the test, 25 human-drawn images and 25 Standard Diffusion generated images were used from the AI-ArtBench test set. The test initiates with a choice question in Figure \ref{fig:turingtest-initial} to understand the participant's knowledge of AI art and Human art. Then, images were displayed in a list, and the respondents were asked to guess the origin of the image, as in Figure \ref{fig:turingtest-qs}. The images were selected randomly through all ten art style classes and shuffled randomly in each response collection. A 20-minute time limit was imposed per response session to overcome any unfair advantages. The test was shared only among a selected group to ensure the reliability of the results, collecting 50 responses. All responses were collected anonymously. Then, the same image set was used in the ArtBrain AI application.

\begin{table}[]
\centering
\caption{‘Artistic Turing Test’ Results categorised under respondents’ knowledge levels. The response count is shown within brackets in each cell.}
\label{tab:artistic-turing-test}
\resizebox{0.75\textwidth}{!}{%
\begin{tabular}{lllll}
\cline{2-5}
 &
  \multicolumn{4}{l}{AI-Art Detection Accuracy of Human} \\ 
  \hline
\backslashbox{Human Art Knowledge}{AI Art Knowledge}&
  Novice &
  Beginner &
  Advanced &
  Expert \\ \hline
\begin{tabular}[c]{@{}l@{}}Novice\\ (Response Count)\end{tabular} &
  \begin{tabular}[c]{@{}l@{}}50.0\%\\ (12)\end{tabular} &
  \begin{tabular}[c]{@{}l@{}}58.8\%\\ (8)\end{tabular} &
  \begin{tabular}[c]{@{}l@{}}70.0\%\\ (1)\end{tabular} &
  \begin{tabular}[c]{@{}l@{}}-\\ (0)\end{tabular} \\
\begin{tabular}[c]{@{}l@{}}Beginner\\ (Response Count)\end{tabular} &
  \begin{tabular}[c]{@{}l@{}}45.5\%\\ (4)\end{tabular} &
  \begin{tabular}[c]{@{}l@{}}51.4\%\\ (16)\end{tabular} &
  \begin{tabular}[c]{@{}l@{}}62.0\%\\ (8)\end{tabular} &
  \begin{tabular}[c]{@{}l@{}}-\\ (0)\end{tabular} \\
\begin{tabular}[c]{@{}l@{}}Advanced\\ (Response Count)\end{tabular} &
  \begin{tabular}[c]{@{}l@{}}52.0\%\\ (2)\end{tabular} &
  \begin{tabular}[c]{@{}l@{}}55.0\%\\ (2)\end{tabular} &
  \begin{tabular}[c]{@{}l@{}}-\\ (0)\end{tabular} &
  \begin{tabular}[c]{@{}l@{}}-\\ (0)\end{tabular} \\
\begin{tabular}[c]{@{}l@{}}Expert\\ (Response Count)\end{tabular} &
  \begin{tabular}[c]{@{}l@{}}-\\ (0)\end{tabular} &
  \begin{tabular}[c]{@{}l@{}}-\\ (0)\end{tabular} &
  \begin{tabular}[c]{@{}l@{}}80.0\%\\ (1)\end{tabular} &
  \begin{tabular}[c]{@{}l@{}}-\\ (0)\end{tabular} \\ \hline
\textbf{\begin{tabular}[c]{@{}l@{}}Overall Human Accuracy\\ (Response Count)\end{tabular}} &
  \multicolumn{4}{l}{\textbf{\begin{tabular}[c]{@{}l@{}}53.8\%\\ (50)\end{tabular}}} \\ \hline
\textbf{ArtBrain (AI) Accuracy} &
  \multicolumn{4}{l}{\textbf{98\%}} \\ \hline
\end{tabular}%
}
\end{table}

\begin{figure}[]
    \centering
    \includegraphics[width=0.75\textwidth]{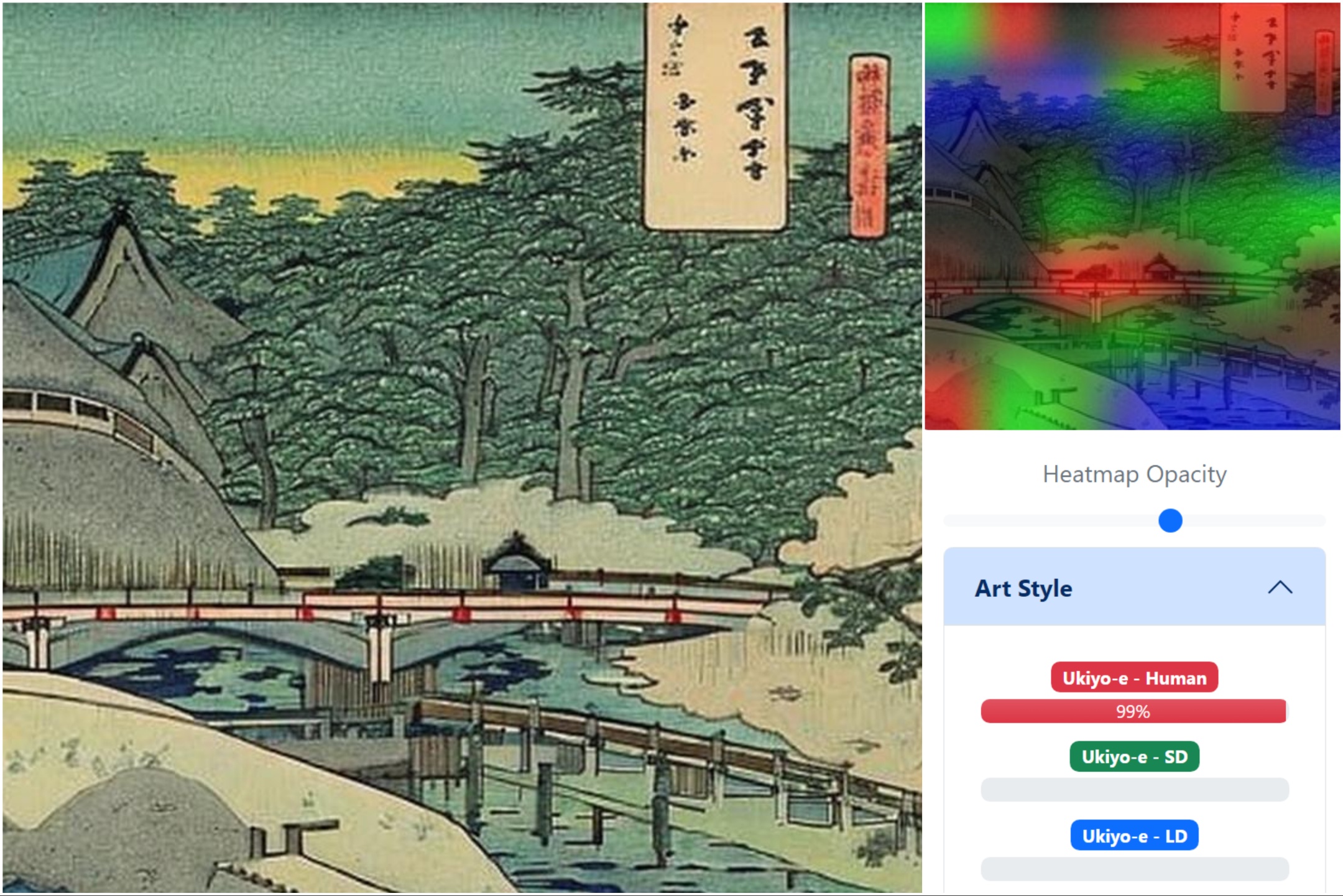}
    \caption{The only incorrect prediction made by the ArtBrain model and its accompanying saliency map. The model wrongly predicted the art as a Human art.}
    \label{fig:artbrainlow-acc}
\end{figure}

Table \ref{tab:artistic-turing-test} represents the results of the conducted Turing test in a matrix form grouping on the respondents' knowledge levels. The maximum accuracy reached is 80\%. However, that is by an expert respondent. The overall accuracy of the respondents is around 54\%, slightly higher than the random chance, which is in this case, 50\%. AI got 98\% on the same set of images, detecting only one AI art wrongly as a Human art, as in Figure \ref{fig:artbrainlow-acc}. The explainable AI heatmap shows that mostly the Japanese-like letters in the image have deceived the ArtBrain model into thinking it is human art. Also, note that ArtBrain's AI model has never been trained on any of these images since they were taken from the test set of the AI-ArtBench dataset. Figure \ref{fig:turingtest-qs} shows the art, which only 16\% of respondents correctly identified as `machine-generated'. 

The results presented in Table \ref{tab:artistic-turing-test} show that the accuracy increases with the knowledge of AI-generated art. In contrast, the accuracy remains low for respondents with advanced knowledge of human art but with novice or beginner knowledge of AI art. This implies that knowledge regarding AI-generated art is more effective and influential in identifying AI-human art than Human art knowledge.

\section{Discussion}\label{discussion}

This chapter drives an in-depth discussion regarding notable aspects of this research, including the authors' opinions based on the evaluation results.

\paragraph{Lack of AI Art}\label{lack-of-ai-art}

One of the key obstacles faced in initiating this research project is the lack of reliable data sources. Virtually no published studies were found to be done on understanding the nature and investigating the issues behind AI art generation. Also, there were no AI-generated artistic datasets available. To overcome this, the study presents AI-ArtBench, a sufficiently sized dataset with more than 120,000 AI-generated artworks and 60,000 real art images from the ArtBench dataset, totalling 180,000+ images altogether (see Table \ref{tab:dataset-info}). However, the dataset can be further improved by adding more modifiers to the generation prompts. In this dataset, only the style was used in the generation prompt in image generation focusing on the capability of diffusion models to generate art based on a given style and the possibility of backtracking their source.

\paragraph{Accuracy of ArtBrain}\label{accuracy-of-artbrain}

As shown in Section \ref{sec:evaluation}, the accuracy of the trained model is very reliable. The human-drawn art classification accuracy surpasses the current state-of-the-art. The attribution accuracy levels are also high, implying that there is a distinct separation in the data distributions from the three sources that human eye cannot perceive. This confirms the findings of Sha et al. \cite{sha_-fake_2022} regarding the idea of ``fingerprint'' that exists in the data distributions, making it possible to attribute the data to its original source. However, the study \cite{sha_-fake_2022} was done on general images and this research, ArtBrain, proves its relevance to AI art images.

\paragraph{Grad-CAM Vs Author's Approach}\label{grad-cam-vs-authors-approach}

\begin{figure}[]
    \centering
    \includegraphics[width=0.75\textwidth]{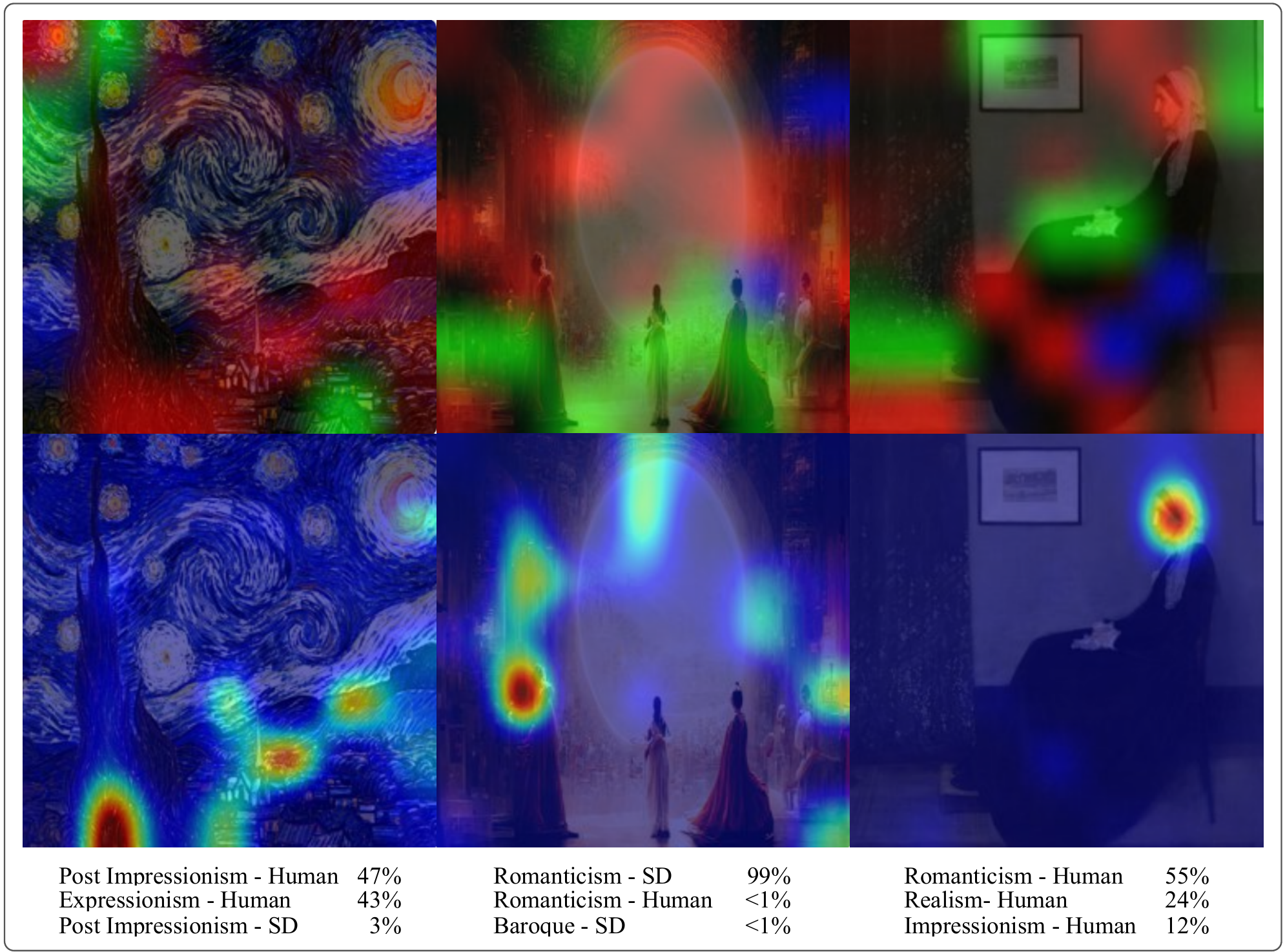}
    \caption{Grad-CAM Vs. FM-G-CAM Comparison.}
    \label{fig:fmcamg-comparison}
\end{figure}

This study uses FM-G-CAM \cite{silva2023fmgcam} to improve the Explainability of CNN predictions. Currently, the prevalent approach is Gradient-weighted Class Activation Map - Grad-CAM \cite{selvaraju_grad-cam_2017}. However, it was identified that even though Grad-CAM is better at localising the objects that may have had an influence on the final prediction, the information it provides is not useful when there are spread-out visual structures that cause the prediction outcome. This is the case for ArtBrain as well since art images are not classified based on a contained object(s) in the image but by the multiple complex visual structures and colour palettes. This study utilises the Fused Multi-class Gradient-weighted Class Activation Map - FM-G-CAM to overcome this issue. As visualised in Figure \ref{fig:fmcamg-comparison}, FM-G-CAM produces a multi-colour map corresponding to multiple classes. Unlike Grad-CAM, FM-G-CAM normalises the weighted activation maps before passing into a final ReLU activation, making low activations visually apparent.

Authors Collins et al. \cite{collins_deep_2018} present a similar method to visualise multiple classes using matrix factorisation; however, it is also focused on object detection. As shown in Figure \ref{fig:fmcamg-comparison}, FM-G-CAM brings out more information from the activations created by the input image without focusing solely on the most important area, as in Grad-CAM. Further, the new approach explains all three top predicted classes, giving a complete explanative visual map to the displayed results. Heatmaps in the right corner of Figure \ref{fig:fmcamg-comparison} further demonstrate the importance of a multi-class map since the anticipated style is realism and the predicted style is romanticism. Even though the predicted top class is inaccurate, the FM-G-CAM heatmap helps to understand why the image can also be classified as a realism art (pointing out human realism features like hands). In contrast, the Grad-CAM heatmap only shows the top predicted class based on a minimal, contained area on the art.

\paragraph{External Factors that Affect the Prediction
Results}\label{external-factors-that-affect-the-prediction-results}

As shown in Section \ref{sec:quantitative-eval}, many external factors can strongly affect the prediction results. This must also be thoroughly researched since most artworks are drawn traditionally using artistic brushes on physical paper, and it must be converted into a digital format to be used on a model. The results of the quantitative evaluation show that they can significantly change the prediction result. Tunable digital image parameters like `Contrast' are good examples for this.

Further, the capturing or scanning device can also affect the prediction results. As shown in preceding Section \ref{predictions-on-replicated-images}, even though it seems identical to the human vision, the model `sees' it in a significantly different way, inverting the initial prediction results. Scanning parameters like DPI (dots per inch) and colour sensitiveness may also affect the prediction results. Malicious manipulation of effects like this can be used to trick the detection system.

\section{Conclusion and Future Work}\label{conclusion}

Considering all the results presented in this report, especially the `Artistic Turing Test' in Section \ref{artistic-turing-test}, it can be confirmed that AI is more effective and accurate than humans in detecting AI Art. However, this research only investigates this hypothesis using diffusion-based models and CNNs. The hypothesis statement can be updated as it pertains to the results of this research as follows: \emph{AI diffusion-generated artworks are better identified and described by CNNs than by humans}.

Further, \emph{the study provides a potentially generalisable approach for verifying the source of AI-generated images}. Even though this method should be tested further to scientifically verify its generalisability, the initial tests presented in Section \ref{sec:quantitative-eval} on popular artworks point towards a positive direction. Also, this opens a new research pathway in developing AI systems to detect artefacts generated using AI. Following are the main limitations identified in the study and the improvements that can be made to overcome each one of them.

Lack of variety and quantity in the dataset samples:
More artistic styles and various generation prompts should be used to overcome this. Real art samples should also be collected, tallying with the generated images. The results of ablation tests in Section \ref{ablation-tests} imply the lack of training data and how transfer learning helps to mitigate that drawback. 

Limited responses and images of the ‘Artistic Turing Test’: The test should be carried out further to collect more responses from a variety of backgrounds. Furthermore, more images should be used randomly, replacing the initial images and increasing the effective number of images used in the overall survey. Additionally, the resolution, quality and factors like DPI should be considered when selecting the images since most of the human artworks are scanned, and AI art is originally digital and hence looks crisper.

Lack of exploitation protection mechanism: The possibility of using the system to trick the system could be mitigated by limiting the attempts of predictions from a single user. Also, a mechanism can be developed to repeatedly limit the prediction of the same/altered image(s).

Limited generative models were used in data generation: The variety of the dataset should be improved by using more types of diffusion models (like very recent Stable Diffusion XL) for the image generation. This will further verify the attribution accuracy scores.

Based on the findings presented in the preceding sections, this research will be a part of a broader attempt to make AI systems more responsible, explainable, and identifiable, ensuring the authenticity of human creations.

\section{Data Availability Statement}
The dataset generated as a part of this study is openly available at Kaggle\footnote{AI-ArtBench dataset available here: \url{https://www.kaggle.com/datasets/ravidussilva/real-ai-art}} under the MIT licence. The Python code written in this study including the model code is available open-source via GitHub: \url{https://github.com/SuienS/ai-art-detector}.

\bibliographystyle{elsarticle-num} 
\bibliography{cas-refs}





\newpage
\appendix
\renewcommand{\thefigure}{A-\arabic{figure}}
\setcounter{figure}{0}

\section{ArtBrain predictions on popular artworks.}
\begin{figure}[h]
    \centering
    \includegraphics[width=0.6\textwidth]{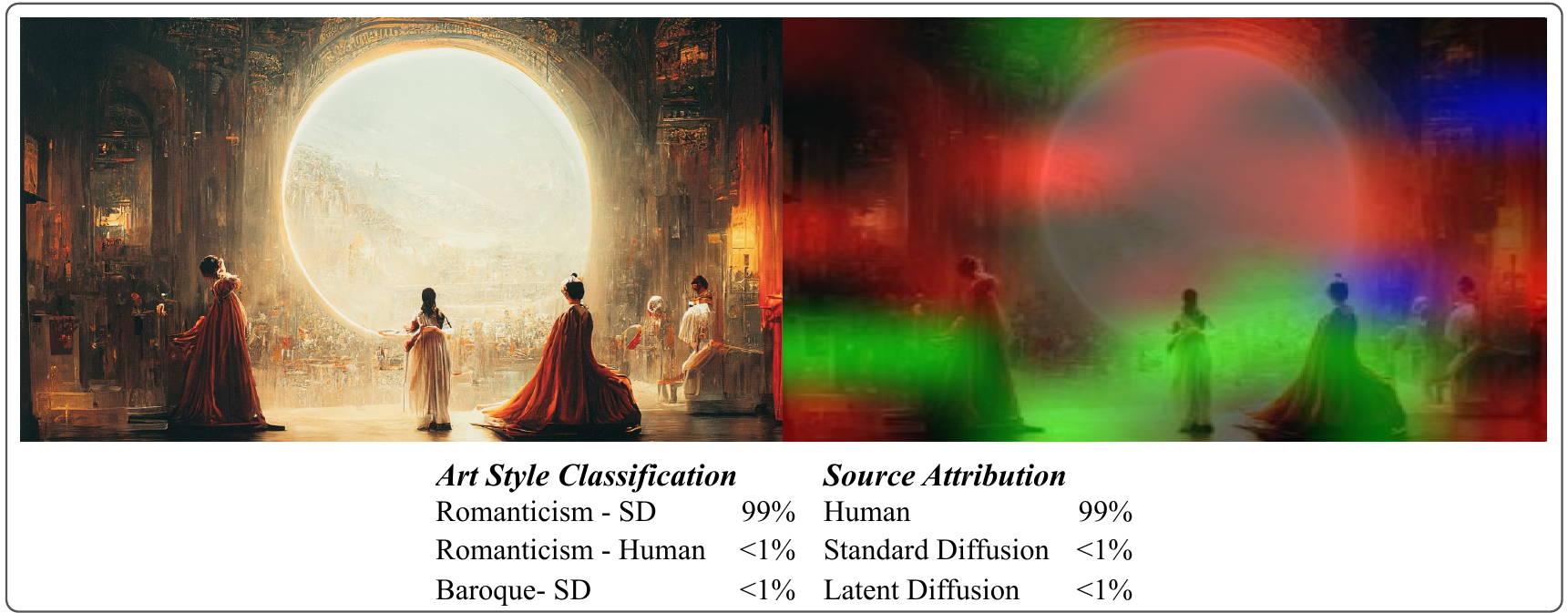}
    \caption{ArtBrain prediction on ``Space Opera" AI Art by Jason Michael Allen.}
    \label{fig:spaceOpera-pred}
\end{figure}

Following Figures \ref{fig:monalisa-pred}, \ref{fig:starrynight-pred}, and \ref{fig:spaceOpera-pred} shows the ArtBrain model's prediction results of a few selected Human and AI-generated artworks.

\begin{figure}[h]
    \centering
    \includegraphics[width=0.5\textwidth]{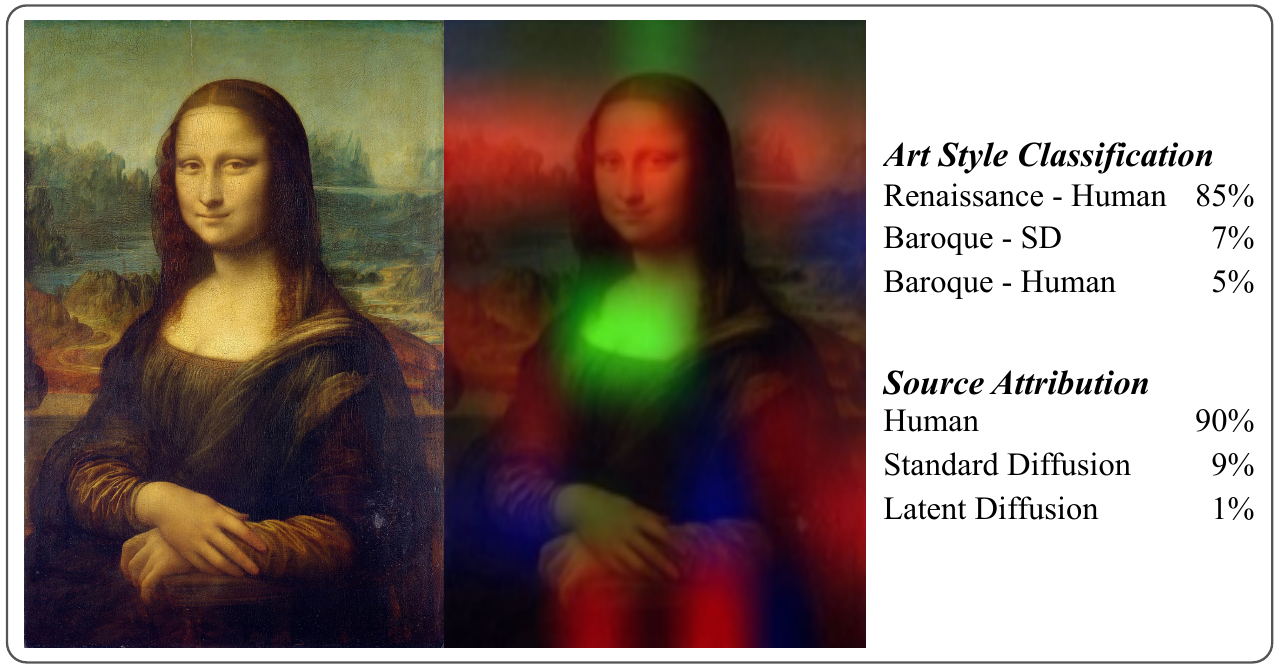}
    \caption{ArtBrain prediction on ``Mona Lisa" by Leonardo Da Vinci.}
    \label{fig:monalisa-pred}
\end{figure}
\vspace{-1em}
\begin{figure}[h]
    \centering
    \includegraphics[width=0.6\textwidth]{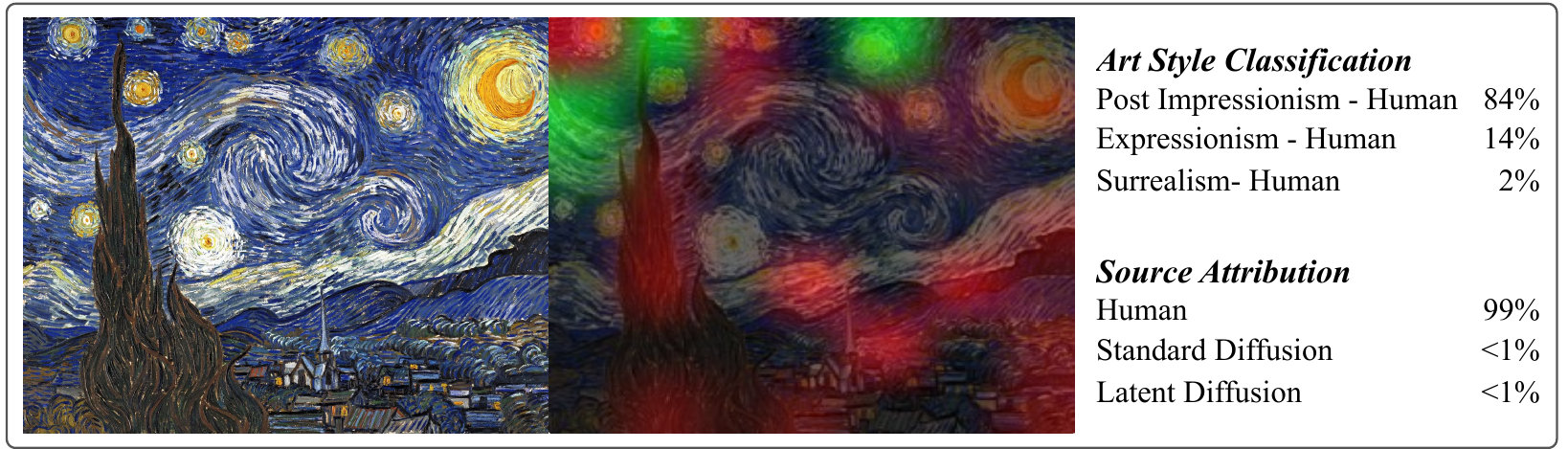}
    \caption{ArtBrain prediction on ``Starry Night" by Vincent Van Gogh.}
    \label{fig:starrynight-pred}
\end{figure}
\vspace{-1em}

\end{document}